\documentclass[letterpaper, 10 pt, conference]{ieeeconf}  %

\IEEEoverridecommandlockouts                              %

\usepackage{amsmath}
\usepackage{amssymb}
\usepackage{bm}
\usepackage{graphicx}
\usepackage{subcaption}
\usepackage{booktabs}
\usepackage{multirow}
\usepackage{arydshln}  %
\newcommand{\hdashlinesep}{\hdashline\addlinespace[2pt]}

\usepackage{url}

\makeatletter
\def\subsubsection{\@startsection{subsubsection}{3}{\parindent}%
  {0ex plus 0.1ex minus 0.1ex}{0ex}{\normalfont\normalsize\bfseries}}
\makeatother

\newcommand{\rqhead}[2]{\par\smallskip\noindent\textbf{#1} #2\par\nobreak}

\title{\LARGE \bf
Smooth Action Chunking via Movement-Primitive Representations \\ for Imitation Learning}

  \author{Shota Nomura$^{1,2}$, Takayuki Osa$^{2}$, Yasuhiro Kato$^{1}$, Kazuki Ota$^{1,2}$, Naoki Shitanda$^{1,2}$, Tatsuya Harada$^{1,2}$%
  \thanks{$^{1}$The University of Tokyo, Tokyo, Japan.
          {\tt\small \{nomura, kato, ota, shitanda, harada\}@mi.t.u-tokyo.ac.jp}}%
  \thanks{$^{2}$RIKEN Center for Advanced Intelligence Project, Tokyo, Japan.
          {\tt\small takayuki.osa@riken.jp}}%
  }

\begin{document}
\maketitle
\thispagestyle{empty}
\pagestyle{empty}

\begin{abstract}
Imitation learning has become a widely adopted approach for vision-based robot policies, owing to its conceptual simplicity, scalability, and ability to leverage large-scale datasets. As these policies have grown more capable, action chunking has become their standard interface. At each decision step the policy outputs several future actions at once. Backbones from diffusion policies to large vision-language-action models all share it.
Whatever backbone produces the chunk, its temporal structure is learned only implicitly. The commanded trajectory is therefore jerky both within a chunk and between successive chunks. On real hardware this appears as visible vibration and as repeated saturation of the follower's velocity limit.
We address this issue by incorporating a movement-primitive-based trajectory representation into action chunking. Specifically, a trajectory segment is represented as a linear combination of smooth basis functions, yielding smooth trajectories by construction while leaving the policy architecture, training objective, and inference procedure unchanged.
The same representation also yields a closed-form, inference-time correction for the second source of jerk. We condition the predicted coefficients so that the new chunk passes through a via-point at the previous chunk's last commanded action. This correction needs no retraining and, unlike asynchronous Real-Time Chunking, applies to policies queried in one shot.
We instantiate it on a one-shot (ACT) and an iterative (DiT) backbone and evaluate in simulation and on a real bimanual ALOHA platform.
The representation lowers within-chunk jerk by an order of magnitude on the one-shot backbone and by $2.4\times$ on the iterative one, with task success statistically non-inferior. The conditioning then removes $73$--$84\%$ of the boundary discontinuity, at negligible per-step cost.
\end{abstract}

\section{Introduction}

\begin{figure}[t]
  \centering
  \includegraphics[width=\linewidth]{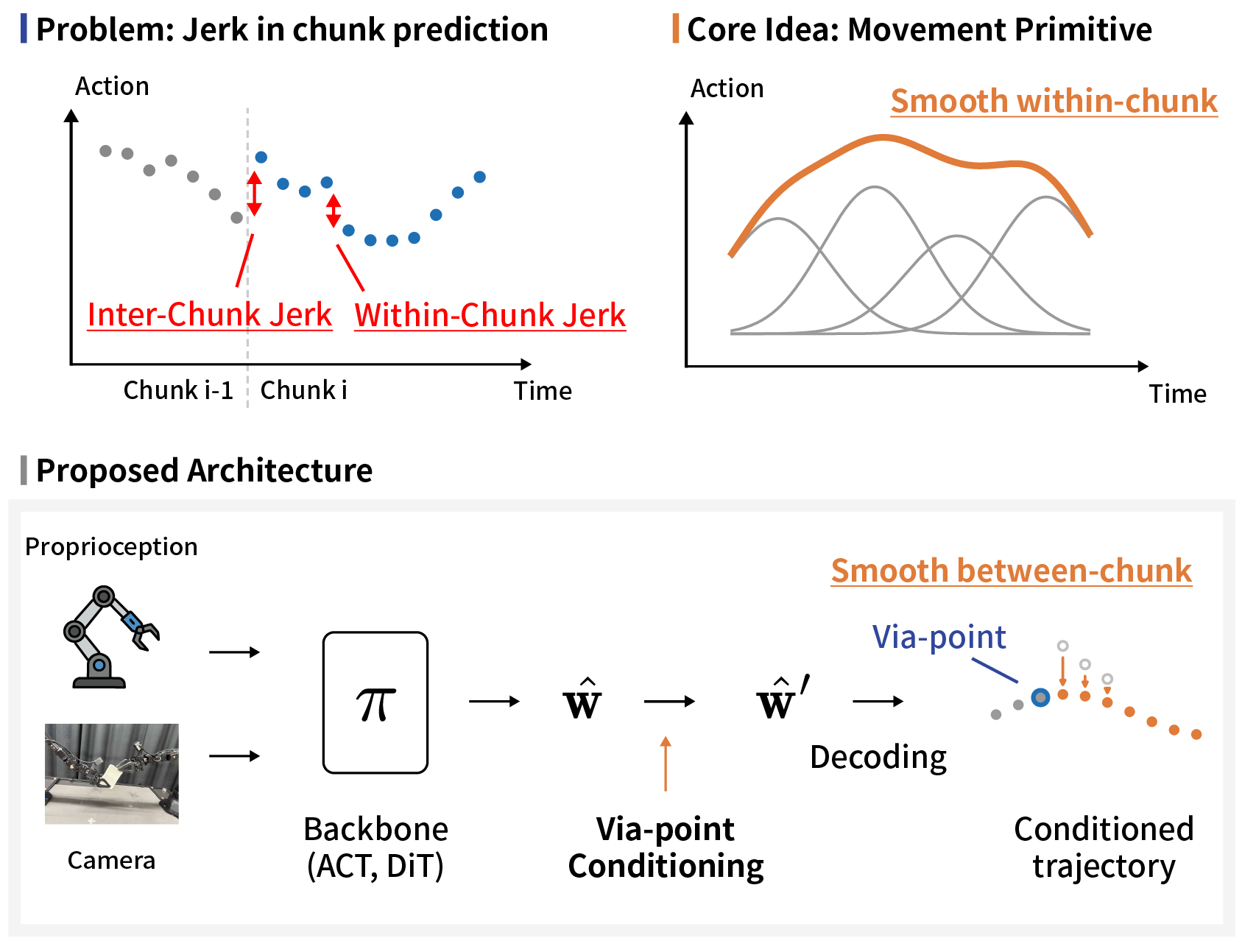}
  \caption{Overview of this study. \textbf{Top left:} predicting the raw action chunk leaves the commanded trajectory jerky within a chunk and discontinuous between chunks. \textbf{Top right:} predicting it as a weighted sum of smooth basis functions is smooth by construction. \textbf{Bottom:} the backbone predicts movement-primitive coefficients $\hat{\boldsymbol w}$. At inference we condition them on the previous chunk's last point to obtain the corrected coefficients $\hat{\boldsymbol w}'$, then decode these into the action chunk.}
  \label{fig:overview}
\end{figure}

Imitation learning has become a widely used route to robot policies that generate continuous actions in real time from visual observations, without explicitly modeling the environment or the task.
As these policies handle a broader range of tasks, \emph{action chunking}, in which the policy outputs a segment of several future actions at once rather than a single action per step, has become their standard interface.
It realizes forward-looking behavior and is beneficial both for task performance and for reducing the frequency at which the policy must be queried.
The interface is shared across otherwise very different designs, from lightweight imitation-learning policies that regress a chunk in one shot, to diffusion-based policies that produce it through an iterative denoising loop, to the large vision-language-action (VLA) models increasingly used for robot control~\cite{pi05,nvidia2025gr00tn1openfoundation}.

Typically, the temporal structure of action chunks is learned only implicitly, so the generated trajectories are often jerky. On real hardware this manifests as noticeable vibration and frequent saturation of the velocity limits of the follower arm, which tracks the commanded joint positions. Such vibration hinders rapid task execution and can accelerate the mechanical fatigue of the robot arm.

This jerk has two distinct sources.
Within a chunk, neighboring predicted steps are not constrained to vary smoothly.
Between successive chunks, the start of a newly predicted chunk need not align with the tail of the chunk currently being executed, producing a discontinuity at every chunk boundary.
Existing remedies are largely external, such as temporal ensembling, low-pass filtering, or Real-Time Chunking (RTC)~\cite{NEURIPS2025_300ccb21}.

We address both sources of jerk by predicting trajectories in the coefficient space of a movement-primitive (MP) trajectory representation, through two components of different scope, each targeting one source.
First, to remove the within-chunk jerk, we adopt a smooth trajectory representation, in which a chunk is predicted as a linear combination of smooth basis functions, so that every chunk is smooth by construction.
This is purely a change of prediction target and applies to any chunk-based backbone.
Second, to remove the between-chunk discontinuity, we add an inter-chunk conditioning step. At inference time we condition the predicted chunk on \emph{via-points}, anchor points taken from the previous chunk's last commanded actions. This uses the closed-form conditioning update of Probabilistic Movement Primitives (ProMP)~\cite{paraschos2013promp}, so that the new chunk starts close to where the previous one ended.
The correction requires no additional training. Because it acts on a single predicted chunk rather than on a denoising loop, it reaches one-shot policies that RTC-style asynchronous stitching cannot, while remaining applicable to iterative ones (Fig.~\ref{fig:overview}).

Both components are independent of the policy architecture, applying whether the chunk is produced in one shot or through an iterative loop. We therefore instantiate them on two backbones at opposite ends of that axis, the lightweight one-shot Action Chunking with Transformers (ACT)~\cite{zhao2023act} and an iterative Diffusion Transformer (DiT)~\cite{11127493}, and evaluate both in simulation and on a real bimanual ALOHA platform.
We make the following contributions:
\begin{itemize}
  \item \textbf{A smooth trajectory representation.} Predicting a chunk as a combination of smooth basis functions removes within-chunk jerk by construction, leaving the backbone, its objective, and its inference procedure unchanged.
  \item \textbf{A training-free inter-chunk correction.} Inference-time ProMP conditioning on via-points from the previous chunk removes most of the boundary discontinuity and applies even to one-shot policies.
  \item \textbf{Empirical validation.} Across both backbones, in simulation and on a real bimanual ALOHA platform, the method improves trajectory smoothness while keeping task success comparable to the baselines.
\end{itemize}

\section{Related Work}

\paragraph{Chunk-based imitation learning}
Predicting a chunk of future actions has become a common design choice in imitation learning~\cite{Osa18a,doi:10.1177/02783649261417694}.
ACT~\cite{zhao2023act} predicts a fixed-horizon action chunk with a CVAE~\cite{Kingma14,Sohn15} and blends overlapping chunks with temporal ensembling, while Diffusion Policy~\cite{chi2023diffusionpolicy} generates the chunk through a denoising diffusion process.
Diffusion Transformer policies~\cite{11127493} scale this idea with a transformer backbone and improved training stability.
The same chunking mechanism underlies most recent large vision-language-action models~\cite{pi05,nvidia2025gr00tn1openfoundation}.

\paragraph{Trajectory smoothness}
Jerky commanded trajectories are a known failure mode of chunk-based policies and are typically mitigated externally, e.g.\ by temporal ensembling, low-pass filtering, smoothness-regularized training objectives~\cite{9561138}, or the smoothing behavior of the low-level controller.
Movement-primitive methods such as Dynamic Movement Primitives (DMPs)~\cite{Ijspeert2001,6797340}, Probabilistic Movement Primitives (ProMPs)~\cite{paraschos2013promp}, and ProDMPs~\cite{li2023prodmp} instead build smoothness into the trajectory representation, encoding it with smooth basis functions. We employ a simplified variant of ProMP, in which trajectories are linear combinations of smooth basis functions. ProMP additionally supports probabilistic conditioning on via-points, which we exploit for the inter-chunk connection (Section~\ref{subsec:method_conn}).

\paragraph{Chunk stitching and asynchronous execution}
A separate line of work targets the discontinuity between successive chunks.
The simplest approach blends overlapping predictions, by temporal ensembling~\cite{zhao2023act} or weighted-sum blending, which reduces jumps but lags the commanded trajectory and cannot enforce continuity exactly.
Real-Time Chunking (RTC)~\cite{NEURIPS2025_300ccb21} instead constrains the next chunk to continue the already-committed prefix by soft-inpainting that prefix at every step of the iterative denoising process. By construction this presupposes an iterative policy and an asynchronous schedule, so a one-shot policy, with no denoising loop to inpaint into, lies outside its scope.
Others build both the smoothness and the connection into the action representation itself. ProDMP~\cite{li2023prodmp} adds structural terms that force every decoded trajectory to start from the current position and velocity; BEAST~\cite{beast2025} encodes chunks as B-spline tokens that are smooth within a chunk and continuous across chunk boundaries; and Movement Primitive Diffusion~\cite{10480552}, the closest work to ours, pairs a ProDMP representation with a diffusion policy for gentle manipulation of deformable objects. These bake the connection into a representation the policy is trained to predict, and the diffusion variants tie it to an iterative backbone. Moreover, in our preliminary experiments, policies predicting ProDMP weights fell well short of the action accuracy of their ProMP counterparts on our tasks.
We therefore retain the simpler MP representation as a backbone-agnostic change of prediction target and enforce the connection with a training-free, inference-time correction on a single predicted chunk.

\section{Preliminaries}
Our method combines a movement-primitive trajectory representation with a chunk-based imitation-learning backbone. This section reviews the three ingredients it builds on, namely the MP representation and the two backbones we instantiate it on, ACT and a DiT policy.

\subsection{ProMP}\label{sec:promp}
Probabilistic Movement Primitives (ProMP)~\cite{paraschos2013promp} represent a trajectory as a linear combination of $n$ basis functions.
Let $\tau \in [0, 1]$ denote the normalized phase variable and $\boldsymbol{\phi}(\tau) \in \mathbb{R}^n$ a vector of Gaussian basis functions evaluated at phase $\tau$.
A scalar trajectory $a(\tau)$ is expressed as
\begin{align}
  a(\tau) = \boldsymbol{\phi}(\tau)^\top \boldsymbol{w}, \qquad \boldsymbol{w} \in \mathbb{R}^n
\end{align}
For a multi-dimensional action of dimension $d_a$, the coefficient matrix $\mathbf{W} \in \mathbb{R}^{n \times d_a}$ stacks per-dimension weights.
Over a chunk horizon of $n_{\text{chunk}}$ steps, the full trajectory is recovered in matrix form as
\begin{align}
  \boldsymbol{a}_{t:t+n_{\text{chunk}}} = \boldsymbol{\Psi} \mathbf{W}, \qquad
  \boldsymbol{\Psi} \in \mathbb{R}^{n_{\text{chunk}} \times n}
\end{align}
where row $j$ of $\boldsymbol{\Psi}$ is $\boldsymbol{\phi}(\tau_j)^\top$ for the phase $\tau_j$ corresponding to timestep $t+j$.
Because the basis functions are smooth, any trajectory in this representation is smooth by construction.
ProMP further places a Gaussian prior $\boldsymbol{w}\sim\mathcal{N}(\boldsymbol{\mu}_{\boldsymbol{w}},\Sigma_{\boldsymbol{w}})$ on the weights, under which conditioning the trajectory to pass through a set of via-points admits a closed-form Gaussian update of $(\boldsymbol{\mu}_{\boldsymbol{w}},\Sigma_{\boldsymbol{w}})$. We use both the representation and this conditioning update, the latter instantiated in Section~\ref{subsec:method_conn}.

\subsection{Action Chunking Transformer}\label{sec:act}
Action Chunking with Transformers (ACT)~\cite{zhao2023act} is a CVAE-based imitation learning method that predicts a chunk of $n_{\text{chunk}}$ future actions at each timestep.
An encoder $q_\phi$ maps the demonstration action chunk and current observation $o_t$ to a latent variable $z$, and a decoder reconstructs the chunk $\hat{\boldsymbol{a}}_{t:t+n_{\text{chunk}}}=p_\theta(z,o_t)$.
It is trained with the standard CVAE objective, an $\ell_1$ reconstruction loss on the chunk plus a $\beta$-weighted KL term. At inference $z$ is set to the prior mean and decoded.

\subsection{Diffusion Transformer Policy}\label{sec:diffusion}
As a second backbone we use a Diffusion Transformer (DiT) policy~\cite{11127493}, which, like Diffusion Policy~\cite{chi2023diffusionpolicy}, frames action generation as a denoising diffusion process over the action chunk $\boldsymbol{a}_{t:t+n_{\text{chunk}}}$ but uses a transformer denoiser conditioned on the observation and diffusion timestep.
Following DDPM~\cite{ho2020ddpm}, a network $\epsilon_\theta$ is trained with the usual $\ell_2$ objective to predict the noise added to the chunk, and at inference a chunk is generated by iteratively denoising from $\boldsymbol{a}^K \sim \mathcal{N}(\mathbf{0}, \mathbf{I})$ conditioned on $o_t$.

\section{Proposed Method}
Our method addresses the discontinuity of chunk-based action generation from two sides. Within a chunk, the smooth trajectory representation of Section~\ref{subsec:method_repr} makes every generated chunk smooth. Between chunks, the smooth connection of Section~\ref{subsec:method_conn} aligns each new chunk with the tail of the previous one.

\subsection{Smooth Trajectory Representation}\label{subsec:method_repr}
The core idea of our method is to replace direct action chunk prediction with prediction in the MP coefficient space introduced in Section~\ref{sec:promp}.
Given the current observation $o_t$ (DiT additionally receives $o_{t-1}$), the model outputs $\hat{\boldsymbol{w}}_t \in \mathbb{R}^{n \times d_a}$ instead of $\hat{\boldsymbol{a}}_{t:t+n_{\text{chunk}}}$ directly.
The predicted action chunk is then recovered via $\hat{\boldsymbol{a}}_{t:t+n_{\text{chunk}}} = \boldsymbol{\Psi} \hat{\boldsymbol{w}}_t$.

\paragraph{ACT with the MP representation}
We apply this substitution to ACT by changing the decoder target from the raw action chunk to the MP coefficients.
The decoder $p_\theta$ now predicts $\hat{\boldsymbol{w}}_t$, and the reconstructed action is $\hat{\boldsymbol{a}}_{t:t+n_{\text{chunk}}} = \boldsymbol{\Psi} \hat{\boldsymbol{w}}_t$.
The loss retains the same form as in Section~\ref{sec:act}, with this substitution applied to $\hat{\boldsymbol{a}}$.

\paragraph{DiT with the MP representation}
We apply the same substitution to the DiT policy by replacing $\boldsymbol{a}_{t:t+n_{\text{chunk}}}$ with $\boldsymbol{w}_t$ throughout the noising and denoising steps (Section~\ref{sec:diffusion}).
The loss retains the same form, with the noise-prediction loss evaluated after mapping both the predicted and true noise through $\boldsymbol{\Psi}$, and at inference the trajectory is recovered via $\hat{\boldsymbol{a}}_{t:t+n_{\text{chunk}}} = \boldsymbol{\Psi}\hat{\boldsymbol{w}}_t^{0}$ after iterative denoising.

\subsection{Smooth connection between chunks}\label{subsec:method_conn}
The policy outputs an action chunk conditioned on the observations up to time $t$, with nothing tying it to the actions predicted one step earlier.
It therefore does not generate each chunk in continuity with the previous one, since the start of a new chunk need not align with the tail of the one still being executed, producing a sudden jump in the commanded actions at the boundary.

We address this by conditioning the predicted coefficients $\hat{\boldsymbol{w}}_t$ on via-points drawn from the final $m$ steps of the previous chunk.
Concretely, let $\boldsymbol{y}^* = \{\boldsymbol{a}_{t-m}, \ldots, \boldsymbol{a}_{t-1}\}$ denote the via-points and $\boldsymbol{\Psi}_{\text{via}} \in \mathbb{R}^{m \times n}$ the first $m$ rows of $\boldsymbol{\Psi}$, so that the via-points are placed at the start of the new chunk and execution resumes at row $m$. The update is applied to each action dimension independently.
Applying the ProMP conditioning update (Section~\ref{sec:promp}) yields the corrected coefficient mean
\begin{align}
  \boldsymbol{K} & = \Sigma_{\boldsymbol{w}} \boldsymbol{\Psi}_{\text{via}}^\top \left(\Sigma_y^* + \boldsymbol{\Psi}_{\text{via}} \Sigma_{\boldsymbol{w}} \boldsymbol{\Psi}_{\text{via}}^\top \right)^{-1}                          \\
  \hat{\boldsymbol{w}}'_t & = \hat{\boldsymbol{w}}_t + \boldsymbol{K} \left(\boldsymbol{y}^* - \boldsymbol{\Psi}_{\text{via}} \hat{\boldsymbol{w}}_t\right)
\end{align}
where $\Sigma_{\boldsymbol{w}}$ controls the flexibility of the weight distribution and $\Sigma_y^*$ controls how strictly the via-points are enforced, a smaller $\Sigma_y^*$ pulling the trajectory closer to the via-points (the anchor).
$\Sigma_y^*$ is diagonal, and when $m>1$ the entries belonging to older via-points are relaxed by a factor of ten per step, so the most recent commanded action dominates the correction.
The update is a maximum-a-posteriori correction under a fixed prior $\Sigma_{\boldsymbol{w}}$, not a posterior over a distribution the policy itself predicts. The backbone emits a point estimate $\hat{\boldsymbol{w}}_t$, and $\Sigma_{\boldsymbol{w}}$, relative to $\Sigma_y^*$, sets only how far it may be moved to meet the via-point, giving the corrected trajectory $\hat{\boldsymbol{a}}_{t:t+n_{\text{chunk}}} = \boldsymbol{\Psi} \hat{\boldsymbol{w}}'_t$.
Because $\boldsymbol{K}$ depends only on fixed parameters and the basis matrix, it is precomputed, so the correction runs purely at inference at negligible cost, using the same MP policy with and without it and requiring no additional training.
Details on $\Sigma_{\boldsymbol{w}}$, $\Sigma_y^*$, and $m$ are given in Section~\ref{subsec:setup}.

\section{Experiments}
We ask whether predicting in MP coefficient space makes chunk-based policies smoother without sacrificing task success, and how the inter-chunk conditioning compares to alternative connection strategies.

\subsection{Experimental Setup}\label{subsec:setup}

We evaluate our method in two settings.
In simulation, using a bimanual ALOHA in MuJoCo, we generate $50$ demonstrations per task with a scripted expert policy.
On real hardware, a bimanual ALOHA platform, we collected $100$ demonstrations per task ($50$ for ACT on Fold Cloth) via teleoperation in the same environment used during evaluation.
Both policies observe four camera images and the $14$-dimensional joint state, and output absolute joint positions for both arms including the grippers, z-score normalized per dimension.
We measure commanded jerk in both settings, and evaluate both backbones, ACT and DiT, in each.
Both backbones use a chunk horizon of $n_{\text{chunk}}=50$ steps at a $50\,$Hz control rate and query the policy every $25$ executed steps. In simulation, execution pauses during inference. On hardware, inference runs in the background while the arm continues the current chunk, and the new chunk replaces its remainder on arrival, or is blended with it under +WS. There, all configurations skip the steps of a new chunk that elapsed during inference, except the conditioned chunk, which is executed without skipping them (Section~\ref{sec:discussion}).
Each episode runs a fixed number of steps with no early termination on success, set per task from the length of the demonstrations: $400$ ($8\,$s at $50\,$Hz) in simulation and on Pick-and-Pass, $500$ on Peg Insertion, and $700$ on Fold Cloth. Success is judged from the simulator state by a task-specific criterion in simulation, and by an operator at the end of each episode on hardware.
For the conditioning we use a single position via-point ($m=1$) with $\Sigma_y^*=10^{-1}$ in normalized action units, the operating point selected by the sweep in Section~\ref{subsec:ablation}. $\Sigma_{\boldsymbol{w}}$ is the identity, an isotropic prior in the normalized weight space, so the direction of the correction is determined by the basis functions at the via-point.
Hyperparameters otherwise follow the released implementations, apart from the input resolution ($256\times256$) and the training schedule. ACT is trained for 100k steps with batch size 8 and learning rate $10^{-5}$. DiT is trained with learning rate $10^{-4}$ and cosine decay, for 50k steps with batch size 64 on hardware and 100k steps with batch size 16 in simulation. Within a backbone the compared configurations differ only in the prediction target.

\subsubsection{Models}

For each backbone we compare three configurations, all trained with three random seeds. \textbf{Baseline} predicts the raw action chunk directly, with no MP representation. \textbf{+MP} predicts $n=10$ MP coefficients per action dimension instead. \textbf{+MP+condition} (+Cond in the tables) is the same +MP model with inter-chunk via-point conditioning enabled at inference, so the two MP cells share identical weights and differ only in whether the correction is applied.
The basis functions are Gaussians with centers spaced uniformly in phase and a standard deviation of half the center spacing, normalized to sum to one. Each demonstrated chunk is projected onto these basis functions by ridge regression ($\lambda=10^{-6}$) to obtain the target coefficients, and the loss is computed after mapping both the prediction and its target through $\boldsymbol\Psi$.
Table~\ref{table:real} also reports two connection baselines, both applied at inference. Weighted-sum blending (+WS) ramps linearly from the previous to the new chunk over their remaining overlap and runs on both representations. Real-Time Chunking (+RTC), adapted here to the DDIM sampler with guidance weight $2$, runs on the raw-action DiT policy only, since it inpaints the committed action prefix during denoising and so requires actions as the denoised variable.

\subsubsection{Statistical treatment}
Throughout, error bars denote the standard deviation over the three training seeds. The compared configurations share the same initial states, and the +MP and +MP+condition runs share their trained weights as well. Each comparison therefore rests on the paired difference, whose variation is far smaller than the marginal seed-to-seed spread. Where the pairing runs episode by episode we test success with McNemar's exact test. The jerk effects we read from the paired difference and its magnitude rather than from a null-hypothesis test over three seeds. Task-success comparisons are claims of non-inferiority rather than of improvement.

\subsection{Research Questions and Metrics}\label{subsec:eval}
We organize the evaluation around three questions, each with the metric it rests on.

\rqhead{RQ1 (Performance).}{Does predicting in MP coefficient space preserve task success?}
We report the final task success rate after training, averaged over three training seeds: $100$ episodes per seed in simulation, from a fixed set of initial states shared by the configurations being compared so that they can be tested pairwise, and $10$ episodes per seed on real hardware, where each episode index starts from the same nominal object placement in every configuration and all configurations of a backbone are collected in one session to avoid drift between them.
On hardware, the DiT checkpoint is selected on the validation action MSE, the open-loop action error on held-out demonstrations. Other models use the final checkpoint.

\rqhead{RQ2 (Smoothness).}{How much does the representation reduce jerk, and how does the conditioning compare to alternative connection strategies?}
The primary metric is the jerk of the commanded action stream, the mean absolute third finite difference of the commanded arm-joint trajectory at the $50\,$Hz control rate, in normalized action units per $\mathrm{s}^{3}$.
This signal is generated by the policy and carries no sensor noise, so it is differentiated directly.
Gripper dimensions are excluded throughout, since their open/close transients dominate the all-joint value.
Mirroring the two sources of jerk, we split the commanded samples by their position relative to a re-planning seam, where execution switches to a newly predicted chunk: within-chunk jerk, over chunk-interior samples, isolates the representation, and boundary jerk, over samples within $\pm2$ steps of a seam, isolates the connection.
Commanded smoothness need not survive execution, so we additionally report two execution-side quantities: a band-limited jerk over the $1$--$5\,$Hz vibration band, computed from the recorded joint velocity, and the number of velocity-clamp activations per episode. A velocity clamp fires when the commanded displacement of an arm joint exceeds a per-joint velocity limit, set to $1.3\times$ the training data's per-joint $99.9$th-percentile speed plus $0.15\,$rad/s, and is clipped to it, visible as a stall on the physical arm.
We differentiate the recorded velocity, not the position, because the third derivative of position is dominated by 12-bit encoder quantization noise, which masks any effect of the method.

\rqhead{RQ3 (Computational cost).}{What does the conditioning cost per inference call, against the alternatives?}
We compare the inference time per call of each connection strategy. Our conditioning applies one precomputed gain matrix per chunk, whereas RTC modifies the trajectory at every denoising step.

\subsection{Tasks}

\paragraph{Simulation tasks}
We use two bimanual tasks from the ALOHA simulation suite~\cite{zhao2023act} with scripted-expert demonstrations: Transfer Cube, where the right arm picks up a cube and hands it to the left, and Insertion, where the right arm inserts a peg into a socket held by the left.

\begin{figure*}[t]
  \centering
  \begin{subfigure}[b]{0.19\linewidth}
    \includegraphics[width=\linewidth]{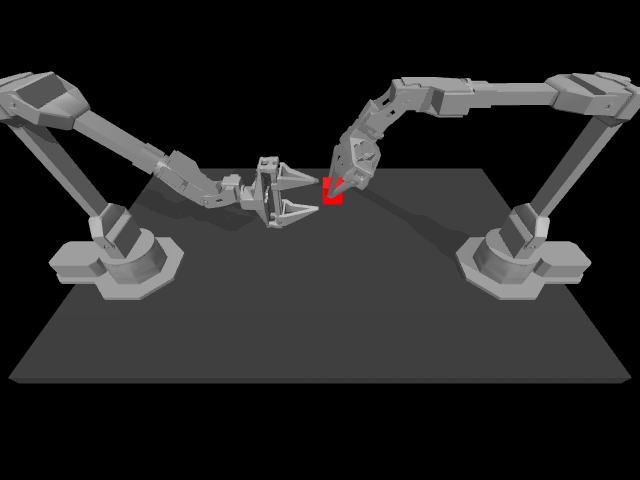}
    \caption{Transfer Cube (sim)}
  \end{subfigure}\hfill
  \begin{subfigure}[b]{0.19\linewidth}
    \includegraphics[width=\linewidth]{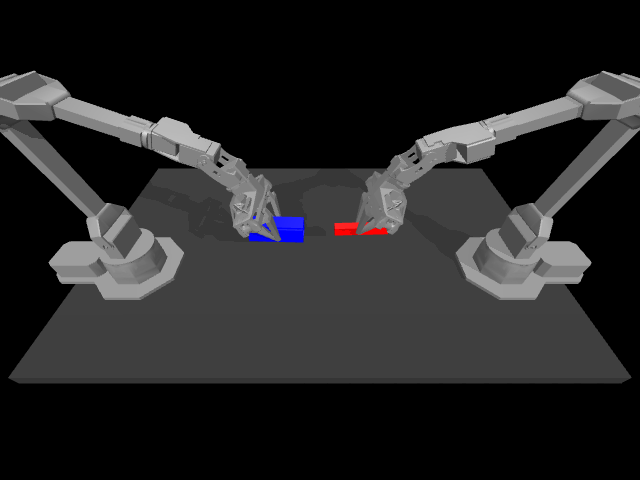}
    \caption{Insertion (sim)}
  \end{subfigure}\hfill
  \begin{subfigure}[b]{0.19\linewidth}
    \includegraphics[width=\linewidth]{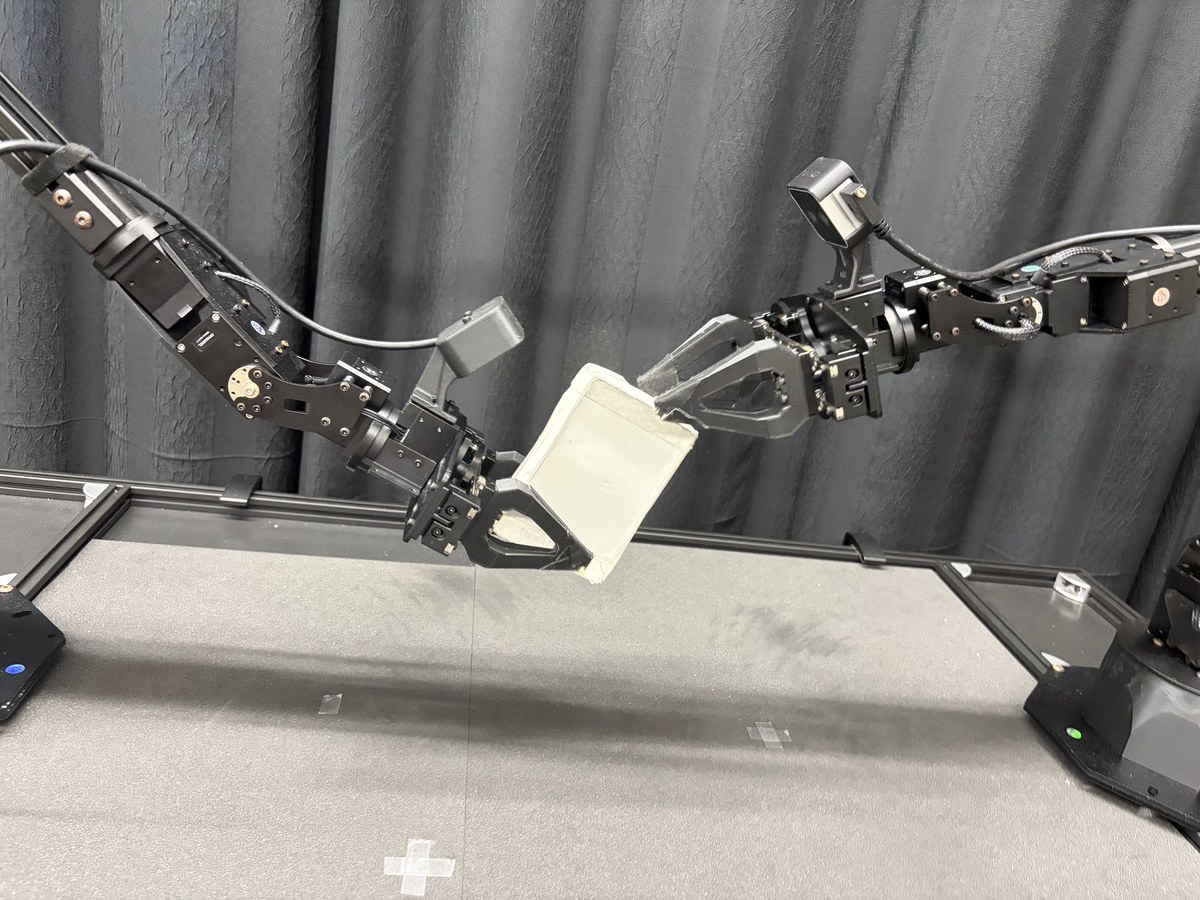}
    \caption{Pick-and-Pass}
  \end{subfigure}\hfill
  \begin{subfigure}[b]{0.19\linewidth}
    \includegraphics[width=\linewidth]{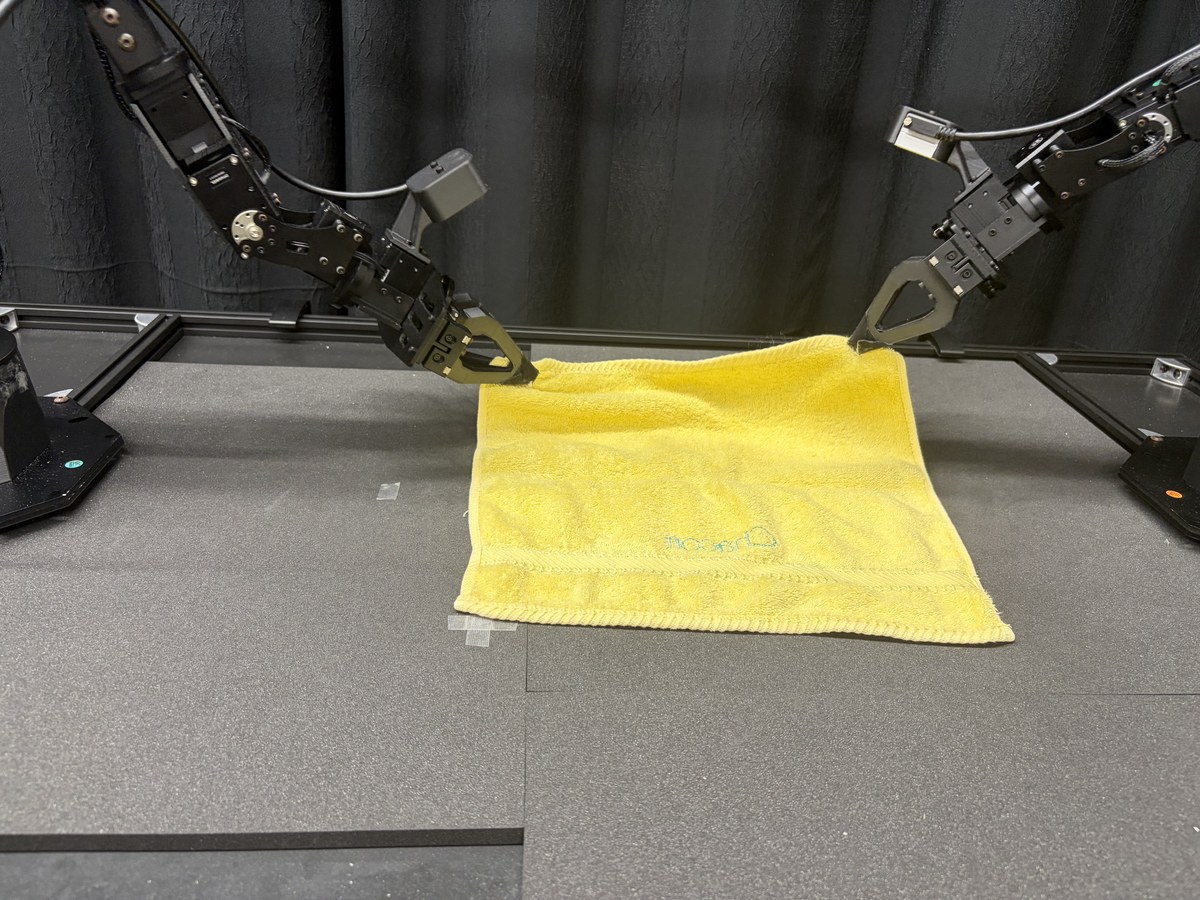}
    \caption{Fold Cloth}
  \end{subfigure}\hfill
  \begin{subfigure}[b]{0.19\linewidth}
    \includegraphics[width=\linewidth]{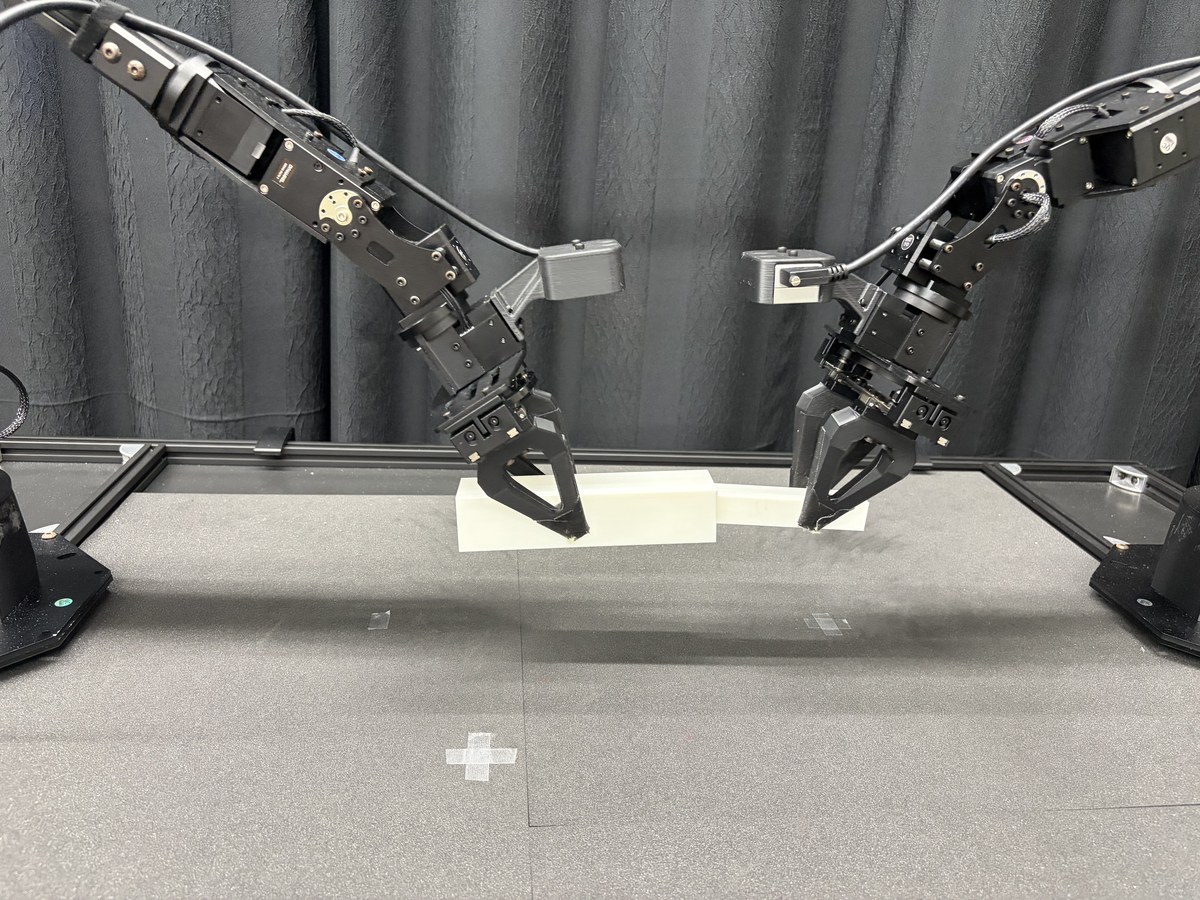}
    \caption{Peg Insertion}
  \end{subfigure}
  \caption{Evaluation tasks. (a, b) Simulation. (c--e) Real ALOHA.}
  \label{fig:tasks}
\end{figure*}

\paragraph{Real-robot tasks}
We evaluate three bimanual tasks on the real ALOHA platform: Pick-and-Pass, where one arm grasps an object and passes it to the other; Fold Cloth, where both arms cooperatively fold a cloth; and Peg Insertion, where one arm holds a socket while the other inserts a peg (Fig.~\ref{fig:tasks}).

\subsection{Results}

Simulation and real-hardware results are reported together under each research question.

\subsubsection{Task Success (RQ1)}\label{subsec:success}
Predicting in MP coefficient space leaves final task success essentially unchanged in both settings.
The fixed initial-state protocol pairs the MP and non-MP runs of a backbone episode by episode, so we compare them with McNemar's test.
In simulation (Table~\ref{table:sim}(a)) final success is flat across the representation in three of the four cells. Only DiT on Transfer Cube moves significantly, and it moves down ($p=0.01$), while adding the boundary conditioning moves no cell ($p\ge0.06$).
On real hardware (Table~\ref{table:real}(a)) the same test, pooling the $30$ index-matched episode pairs of each cell, finds no cell where either the representation or the added conditioning moves success ($p\ge0.45$ throughout), the largest change in any cell amounting to one episode per seed.
With a non-inferiority margin of one episode per seed, the $10$-percentage-point resolution of the $10$-episode protocol, every real cell is non-inferior, and the only exception in either setting is the DiT decrease in simulation above.
The test has limited power, since cells near the ceiling cannot show a drop and the seed-to-seed spread is large relative to the differences at stake.

\begin{table}[t]
  \caption{Simulation results. +Cond is +MP with inter-chunk conditioning. Arrows indicate the better direction, and bold marks the lowest jerk in each row of (b) and (c).}
  \label{table:sim}
  \centering
  \setlength{\tabcolsep}{4pt}
  \begin{tabular}{l l ccc}
    \toprule
    Backbone & Task & Baseline & +MP & +Cond \\
    \midrule
    \multicolumn{5}{l}{\emph{(a) Success rate (\%)~($\uparrow$)}} \\
    \multirow{2}{*}{ACT}
             & Transfer Cube & $96.0\pm5.3$  & $97.7\pm1.2$ & $99.0\pm1.0$ \\
             & Insertion     & $47.7\pm15.8$ & $55.7\pm13.4$ & $54.3\pm6.4$ \\
    \hdashlinesep
    \multirow{2}{*}{DiT}
             & Transfer Cube & $85.7\pm0.6$  & $77.3\pm3.5$ & $78.0\pm6.9$ \\
             & Insertion     & $57.0\pm10.0$ & $55.3\pm5.5$ & $63.0\pm8.9$ \\
    \midrule
    \multicolumn{5}{l}{\emph{(b) Within-chunk commanded jerk~($\downarrow$)}} \\
    \multirow{2}{*}{ACT}
             & Transfer Cube & $261.4\pm10.6$ & $69.0\pm1.6$          & $\mathbf{68.7\pm1.0}$ \\
             & Insertion     & $339.4\pm16.7$ & $62.6\pm1.1$          & $\mathbf{60.7\pm2.8}$ \\
    \hdashlinesep
    \multirow{2}{*}{DiT}
             & Transfer Cube & $907.9\pm931.0$ & $\mathbf{63.5\pm0.8}$ & $70.6\pm3.9$ \\
             & Insertion     & $307.0\pm20.2$  & $\mathbf{59.6\pm2.2}$ & $65.7\pm0.2$ \\
    \midrule
    \multicolumn{5}{l}{\emph{(c) Boundary commanded jerk~($\downarrow$)}} \\
    \multirow{2}{*}{ACT}
             & Transfer Cube & $660\pm83$    & $526\pm77$   & $\mathbf{163\pm5}$ \\
             & Insertion     & $1008\pm118$  & $713\pm62$   & $\mathbf{172\pm3}$ \\
    \hdashlinesep
    \multirow{2}{*}{DiT}
             & Transfer Cube & $1828\pm1021$ & $1874\pm213$ & $\mathbf{305\pm63}$ \\
             & Insertion     & $2200\pm417$  & $2443\pm375$ & $\mathbf{319\pm93}$ \\
    \bottomrule
  \end{tabular}
\end{table}

\subsubsection{Smoothness: Within-Chunk Representation (RQ2)}\label{subsec:repr}
The within-chunk commanded jerk shows the effect of the representation on both platforms.
In simulation (Table~\ref{table:sim}(b)) the MP model is smoother in every cell, and reaches nearly the same level in all four even though the baselines it starts from differ widely.
On hardware (Table~\ref{table:real}(b)) the same signature appears. The representation lowers within-chunk jerk by an order of magnitude on ACT and severalfold on DiT, and on each task the two backbones converge to nearly the same value, a level set by the demonstrated motion the basis must fit.
One simulation seed is an outlier. A single DiT run on Transfer Cube has a far jerkier baseline than the other two seeds at the same success rate, inflating that entry's mean and spread, whereas the +MP entries show no comparable spread anywhere.

\begin{table*}[t]
  \caption{Real-hardware results. Columns are grouped by the representation they run on, and columns within a group share the same trained weights. Arrows indicate the better direction, and bold marks the lowest value in each row of (b)--(e). RTC applies to DiT only.}
  \label{table:real}
  \centering
  \footnotesize
  \setlength{\tabcolsep}{3pt}
  \begin{tabular}{l l ccc ccc}
    \toprule
             &               & \multicolumn{3}{c}{Raw action} & \multicolumn{3}{c}{MP} \\
    \cmidrule(lr){3-5} \cmidrule(lr){6-8}
    Backbone & Task          & Baseline & \; +WS & \; +RTC & +MP & \; +WS & \; +Cond \\
    \midrule
    \multicolumn{8}{l}{\emph{(a) Success rate (\%)~($\uparrow$)}} \\
    \multirow{3}{*}{ACT}
             & Pick-and-Pass & $80.0\pm10.0$ & $83.3\pm20.8$ & -- & $73.3\pm23.1$ & $76.7\pm20.8$ & $83.3\pm15.3$ \\
             & Fold Cloth    & $96.7\pm5.8$  & $96.7\pm5.8$  & -- & $93.3\pm5.8$  & $100.0\pm0.0$ & $100.0\pm0.0$ \\
             & Peg Insertion & $80.0\pm17.3$ & $86.7\pm15.3$ & -- & $80.0\pm10.0$ & $76.7\pm5.8$  & $70.0\pm10.0$ \\
    \hdashlinesep
    \multirow{3}{*}{DiT}
             & Pick-and-Pass & $86.7\pm15.3$ & $96.7\pm5.8$  & $83.3\pm20.8$ & $90.0\pm10.0$ & $96.7\pm5.8$  & $96.7\pm5.8$  \\
             & Fold Cloth    & $96.7\pm5.8$  & $96.7\pm5.8$  & $93.3\pm5.8$  & $90.0\pm0.0$  & $86.7\pm15.3$ & $90.0\pm10.0$ \\
             & Peg Insertion & $90.0\pm10.0$ & $93.3\pm5.8$  & $100.0\pm0.0$ & $96.7\pm5.8$  & $96.7\pm5.8$  & $96.7\pm5.8$  \\
    \midrule
    \multicolumn{8}{l}{\emph{(b) Within-chunk commanded jerk~($\downarrow$)}} \\
    \multirow{3}{*}{ACT}
             & Pick-and-Pass & $280.1\pm10.0$ & $251.2\pm8.7$  & --             & $\mathbf{18.9\pm0.9}$ & $20.5\pm0.9$ & $20.1\pm0.6$ \\
             & Fold Cloth    & $434.2\pm19.1$ & $380.4\pm20.2$ & --             & $\mathbf{33.4\pm1.4}$ & $36.3\pm1.4$ & $34.0\pm1.0$ \\
             & Peg Insertion & $319.5\pm17.6$ & $290.1\pm13.4$ & --             & $\mathbf{20.7\pm0.4}$ & $22.9\pm0.5$ & $21.4\pm0.5$ \\
    \hdashlinesep
    \multirow{3}{*}{DiT}
             & Pick-and-Pass & $59.3\pm1.7$   & $60.7\pm4.2$   & $129.2\pm51.9$ & $\mathbf{22.1\pm3.2}$ & $26.5\pm5.0$ & $27.4\pm2.9$ \\
             & Fold Cloth    & $80.4\pm3.1$   & $78.7\pm2.7$   & $99.7\pm7.3$   & $\mathbf{33.1\pm2.3}$ & $42.0\pm3.8$ & $42.1\pm2.9$ \\
             & Peg Insertion & $71.9\pm8.8$   & $74.7\pm8.1$   & $107.1\pm16.4$ & $\mathbf{20.9\pm1.8}$ & $26.6\pm3.5$ & $26.2\pm2.9$ \\
    \midrule
    \multicolumn{8}{l}{\emph{(c) Boundary commanded jerk~($\downarrow$)}} \\
    \multirow{3}{*}{ACT}
             & Pick-and-Pass & $925\pm22$   & $285\pm3$  & --          & $758\pm171$  & $\mathbf{28\pm3}$  & $127\pm12$ \\
             & Fold Cloth    & $1425\pm103$ & $444\pm16$ & --          & $1222\pm137$ & $\mathbf{45\pm3}$  & $214\pm14$ \\
             & Peg Insertion & $962\pm106$  & $330\pm25$ & --          & $965\pm34$   & $\mathbf{33\pm1}$  & $150\pm11$ \\
    \hdashlinesep
    \multirow{3}{*}{DiT}
             & Pick-and-Pass & $1601\pm446$ & $82\pm10$  & $473\pm268$ & $1404\pm393$ & $\mathbf{54\pm12}$ & $249\pm18$ \\
             & Fold Cloth    & $1667\pm128$ & $108\pm9$  & $501\pm22$  & $1554\pm259$ & $\mathbf{81\pm8}$  & $416\pm56$ \\
             & Peg Insertion & $1052\pm98$  & $96\pm21$  & $388\pm89$  & $1032\pm190$ & $\mathbf{58\pm9}$  & $244\pm34$ \\
    \midrule
    \multicolumn{8}{l}{\emph{(d) Executed $1$--$5\,$Hz band jerk~($\downarrow$)}} \\
    \multirow{3}{*}{ACT}
             & Pick-and-Pass & $7.05\pm0.47$  & $3.79\pm0.13$ & -- & $6.32\pm1.08$ & $\mathbf{3.42\pm0.10}$ & $5.38\pm0.33$ \\
             & Fold Cloth    & $10.54\pm0.50$ & $5.00\pm0.18$ & -- & $9.74\pm0.91$ & $\mathbf{4.57\pm0.13}$ & $8.19\pm0.43$ \\
             & Peg Insertion & $6.77\pm0.88$  & $\mathbf{3.71\pm0.14}$ & -- & $7.48\pm0.27$ & $3.77\pm0.13$ & $5.99\pm0.35$ \\
    \hdashlinesep
    \multirow{3}{*}{DiT}
             & Pick-and-Pass & $10.83\pm2.51$ & $\mathbf{5.25\pm0.96}$ & $5.42\pm0.62$ & $9.98\pm2.76$  & $5.53\pm1.41$ & $7.91\pm0.90$  \\
             & Fold Cloth    & $12.43\pm0.72$ & $\mathbf{6.36\pm0.37}$ & $7.47\pm0.17$ & $12.81\pm2.28$ & $7.79\pm1.13$ & $12.53\pm0.96$ \\
             & Peg Insertion & $7.99\pm0.66$  & $\mathbf{5.04\pm0.43}$ & $5.31\pm0.27$ & $8.88\pm1.88$  & $5.51\pm1.23$ & $7.75\pm0.91$  \\
    \midrule
    \multicolumn{8}{l}{\emph{(e) Velocity-clamp activations per episode~($\downarrow$)}} \\
    \multirow{3}{*}{ACT}
             & Pick-and-Pass & $10.8\pm0.8$ & $2.9\pm0.2$ & -- & $9.7\pm2.9$ & $2.7\pm0.4$ & $\mathbf{2.6\pm0.4}$ \\
             & Fold Cloth    & $12.6\pm2.2$ & $\mathbf{0.9\pm0.3}$ & -- & $11.1\pm2.1$ & $\mathbf{0.9\pm0.1}$ & $\mathbf{0.9\pm0.2}$ \\
             & Peg Insertion & $7.6\pm1.6$ & $\mathbf{2.0\pm0.0}$ & -- & $8.5\pm0.5$ & $\mathbf{2.0\pm0.0}$ & $\mathbf{2.0\pm0.0}$ \\
    \hdashlinesep
    \multirow{3}{*}{DiT}
             & Pick-and-Pass & $24.4\pm7.7$ & $3.3\pm1.3$ & $5.5\pm2.0$ & $22.4\pm7.6$ & $3.6\pm1.0$ & $\mathbf{2.4\pm1.5}$ \\
             & Fold Cloth    & $21.8\pm2.4$ & $\mathbf{1.2\pm0.3}$ & $3.1\pm0.7$ & $21.2\pm6.6$ & $4.1\pm2.0$ & $1.9\pm1.7$ \\
             & Peg Insertion & $11.6\pm2.6$ & $2.8\pm0.5$ & $3.9\pm0.2$ & $10.9\pm4.9$ & $3.5\pm1.5$ & $\mathbf{1.7\pm1.0}$ \\
    \bottomrule
  \end{tabular}
\end{table*}

\subsubsection{Smoothness: Inter-Chunk Connection (RQ2)}\label{subsec:connection}
\begin{figure}[t]
  \centering
  \includegraphics[width=\linewidth]{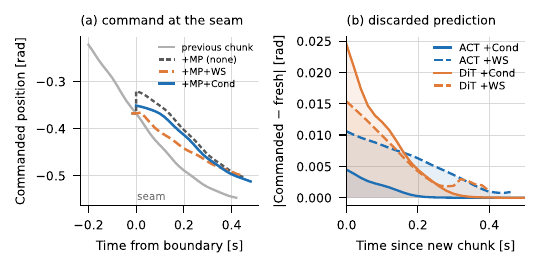}
  \caption{Commanded trajectory around a chunk boundary on Fold Cloth. (a) Command of one representative joint on ACT with no connection, blending, and conditioning. (b) Deviation of the command from the freshly predicted chunk, averaged over arm joints, against time since it arrived, on ACT and DiT. Shading marks the integral.}
  \label{fig:seam}
\end{figure}

Table~\ref{table:real}(c--e) compares four connection strategies, namely no connection, weighted-sum blending, our via-point conditioning, and RTC, on the commanded boundary jerk, the executed $1$--$5\,$Hz band jerk, and velocity-clamp activations. Fig.~\ref{fig:seam} shows the commanded trajectory around a chunk boundary without a connection, with blending, and with conditioning.
Blending is reported on both representations, since it does not depend on how the chunk is parameterized. On the commanded boundary jerk, however, running +WS on top of the MP representation stays well below running it on the raw action in every cell (Table~\ref{table:real}(c)), so on this dimension the representation helps even under blending.
In simulation the conditioning removes most of the boundary jerk in every cell at no cost in success (Table~\ref{table:sim}(c)). The boundary jerk the representation alone leaves is several times larger on DiT than on ACT.

On hardware, the boundary jerk without a connection is tens of times the within-chunk value, and the representation alone reduces it only slightly.
Via-point conditioning removes most of this discontinuity at negligible inference cost (Section~\ref{subsec:cost}) and no change in success, but its effect on the executed motion differs between the two backbones.
On ACT, velocity-clamp activations decrease from about $8$--$13$ per episode without a connection to $1$--$3$ with one on all three tasks (Table~\ref{table:real}(e)), and the executed $1$--$5\,$Hz jerk in Table~\ref{table:real}(d) decreases as well. Blending reduces clamp activations to a similar degree, which suggests that the reduction comes from closing the seam rather than from the way it is closed.
On DiT, clamp activations decrease in the same way, but the executed band jerk stays within the seed spread.
The seam left after conditioning is larger on DiT, about ten times the within-chunk value, and blending, which lowers it further, also lowers the executed band jerk, so the remaining vibration on DiT may be tied to this residual seam.

Three further comparisons concern the other connection strategies. First, RTC reduces the seam by a fraction comparable to our conditioning, at success within noise, while raising within-chunk jerk.
Second, weighted-sum blending attains the lowest boundary and executed band jerk in every cell where it was run, at no measurable inference cost and with success within noise. Applied to the raw-action policy with no MP representation at all, it already reaches a lower executed $1$--$5\,$Hz jerk on ACT than the full representation plus conditioning attains. The two differ in what is commanded to obtain the smoothing (Fig.~\ref{fig:seam}).
Blending emits a convex combination of the previous and the newly predicted chunk across the entire remaining overlap, not merely at the seam, so the command keeps departing from the fresh prediction long after the boundary.
Measuring that departure as the integral of $|\text{commanded}-\text{freshly predicted}|$, conditioning on ACT discards far less of the fresh prediction than blending does ($0.023$ versus $0.128\,$rad$\cdot$step, averaged over arm joints) and returns to it sooner, an advantage that does not hold on DiT ($0.150$ versus $0.131$).
Third, neither connection method substitutes for the representation in the chunk interior. Blending applied to the raw-action baseline leaves within-chunk jerk essentially untouched on every task, worse there than the MP policy with no connection at all.

\subsubsection{Computational Cost}\label{subsec:cost}
We report the inference time per call to address RQ3. Predicting $n$ coefficients instead of $n_{\text{chunk}}$ actions changes only the width of the output head, so we compare against other ways of obtaining a smooth, connected trajectory.
Temporal ensembling in its original form averages all chunks covering the current step and so requires a forward pass per step. At the $50\,$Hz control rate neither backbone's inference on our RTX 6000 Ada GPU fits the $20\,$ms per-step budget (ACT $\approx33\,$ms, DiT $\approx113\,$ms with eight DDIM steps), so we do not report it as a measured arm.
Weighted-sum blending mixes only the overlap between consecutive chunks and our conditioning applies one precomputed closed-form update per chunk. Both leave the time per call indistinguishable from the base policy, within its seed-to-seed spread (DiT $116\,$ms with conditioning versus $113\,$ms without).
RTC, which modifies the trajectory at every denoising step, is the only strategy with a real overhead, nearly doubling the time per call on DiT (from $113$ to $205\,$ms).

\subsection{Ablation Study}\label{subsec:ablation}

We ablate two design choices on real hardware, on Pick-and-Pass with the ACT backbone. The first is the conditioning strength. We sweep the via-point noise $\Sigma_y^*$ over seven orders of magnitude together with the number of via-points $m$. Together they set the operating point of \mbox{+MP+condition}. The second is the number of basis functions, $n\in\{5,10,20\}$, which trades expressiveness against smoothness. The former reuses the trained \mbox{+MP} model at inference, whereas the latter requires a training run per level. Both ablations were run in sessions separate from Table~\ref{table:real}, so their values are compared within each ablation only.

Table~\ref{table:cond} first sweeps the via-point noise $\Sigma_y^*$ with a single via-point ($m=1$) and then varies the number of via-points at the selected $\Sigma_y^*$.
Success does not distinguish the levels, all of which lie within sampling noise of one another.
The commanded boundary jerk falls as $\Sigma_y^*$ is tightened and levels off below $10^{-2}$, whereas the executed jerk is U-shaped, with a reproducible minimum at $10^{-1}$.
We select $\Sigma_y^*=10^{-1}$, the minimum of the executed jerk rather than of the commanded one.
A likely reason for the U shape is that the anchor, the last command of the previous chunk, carries that chunk's own prediction error, which a strict via-point passes into the next chunk and a looser one partly averages out.
Using three or five via-points instead of one leaves the executed jerk unchanged within noise and lowers the commanded boundary jerk only slightly, since the older via-points are relaxed and the most recent one dominates the correction, so we keep $m=1$.

For the number of basis functions, evaluated with two seeds per level, within-chunk commanded jerk rises with $n$ ($4.8$, $19.0$, and $57.1$ for $n=5$, $10$, and $20$), while success varies within the seed spread ($60\pm57$, $75\pm7$, and $65\pm21\%$). We use $n=10$, the middle setting, since a wider basis only adds roughness at no gain in success.

\begin{table}[t]
  \caption{Conditioning ablation on Pick-and-Pass with ACT. All rows use the same \mbox{+MP} model. $\dagger$ marks the selected operating point.}
  \label{table:cond}
  \centering
  \setlength{\tabcolsep}{4pt}
  \begin{tabular}{cc ccc}
    \toprule
    $\Sigma_y^*$ & $m$ & Boundary~($\downarrow$) & Executed $1$--$5\,$Hz~($\downarrow$) & Success (\%)~($\uparrow$) \\
    \midrule
    $10^{-6}$           & 1 & $75\pm3$   & $6.69\pm0.17$          & $83\pm6$ \\
    $10^{-3}$           & 1 & $73\pm2$   & $6.50\pm0.10$          & $83\pm15$ \\
    $10^{-2}$           & 1 & $81\pm4$   & $5.42\pm0.18$          & $87\pm12$ \\
    $10^{-1}$ $\dagger$ & 1 & $121\pm5$  & $\mathbf{4.76\pm0.13}$ & $83\pm15$ \\
    $10^{0}$            & 1 & $398\pm47$ & $5.31\pm0.36$          & $83\pm29$ \\
    $10^{+1}$           & 1 & $740\pm80$ & $6.40\pm0.51$          & $80\pm17$ \\
    \midrule
    $10^{-1}$           & 3 & $108\pm13$ & $4.93\pm0.07$          & $87\pm15$ \\
    $10^{-1}$           & 5 & $99\pm4$   & $4.72\pm0.07$          & $90\pm17$ \\
    \bottomrule
  \end{tabular}
\end{table}

\section{Discussion}\label{sec:discussion}

The two components of our method are not interchangeable.
The representation lowers within-chunk jerk on both backbones regardless of how the coefficients are produced, but barely reduces the seam.
The conditioning removes most of that discontinuity but does not smooth the chunk interior (Section~\ref{subsec:connection}).
A smooth command requires addressing both sources.

\paragraph{Scope: one-shot vs.\ iterative backbones}
Conditioning is backbone-agnostic, but the gap it must close differs between the two backbones.
A fresh DiT chunk starts farther from the anchor than an ACT chunk, about five times farther at the seam (Fig.~\ref{fig:seam}(b)).
Because the correction moves the weights of the basis functions that overlap the via-point, it does not stay at the seam but decays over roughly the first third of the chunk (Fig.~\ref{fig:seam}(b)). On DiT this decay starts from a far larger displacement, which is why conditioning leaves a larger residual seam there, one possible reason the executed vibration band does not improve.

\paragraph{Limitations}
Blending is the stronger method on the boundary and executed jerk metrics (Section~\ref{subsec:connection}).
Two properties of our benchmark, however, keep its costs from showing.
The tasks are quasi-static with a fixed episode length, so the cost of averaging away the fresh prediction, a delayed reaction, does not register in success rate.
The demonstrated motions are also nearly unimodal, so the known failure mode of averaging independently predicted chunks, namely that the average of two valid chunks need not itself be valid~\cite{NEURIPS2025_300ccb21}, cannot show up either.
Independently of the benchmark, blending also presupposes an execution schedule that leaves an overlap to average over, whereas conditioning needs only the tail of the previous chunk.
Conditioning may also slow the reaction. On Pick-and-Pass, the one task where we measured time-to-success, completion was slower with it, slightly on ACT and by about a second on DiT. Part of this may come from our hardware implementation rather than from the conditioning itself, since the conditioned chunk was executed without skipping the inference delay and so lagged the fresh prediction by about two steps on ACT and five to eight on DiT. The executed jerk of DiT with conditioning may be affected for the same reason, in addition to the residual seam noted above.
We used Gaussian basis functions following ProMP, but any smooth basis that is linear in the coefficients, such as B-splines~\cite{beast2025}, can be used in the same way, although we have not tested this.
The number of basis functions also limits how fast the command can change within a chunk. On Pick-and-Pass this did not change success beyond the seed spread for $n\in\{5,10,20\}$ (Section~\ref{subsec:ablation}). Since the policy re-plans every $25$ steps, faster motions could be accommodated by a larger $n$ or a shorter chunk, at the cost of more within-chunk roughness.
Comparisons on tasks with multimodal or faster motion, including perturbation tests, are left for future work.

\section{Conclusion}

We addressed the two distinct sources of jerk in chunk-based policies by predicting trajectories in a movement-primitive coefficient space.
Predicting a chunk as a linear combination of smooth basis functions makes it smooth by construction and sharply reduces within-chunk jerk on both backbones, with task success statistically non-inferior. Conditioning the predicted coefficients on via-points from the previous chunk then removes most of the boundary discontinuity that the representation alone leaves. Unlike asynchronous stitching, the conditioning needs no denoising loop to inpaint into, so it remains available to one-shot policies.
A blending baseline suppresses seam jerk further, but only where chunks overlap, and leaves the interior of a raw-action policy rough.
These results indicate that smoothness in chunk-based control can be obtained through the prediction target itself, without external filtering or averaging.

\addtolength{\textheight}{0cm}

 \section*{ACKNOWLEDGMENT}
 This work was partially supported by JST Moonshot R\&D Grant Number JPMJPS2011 and by the New Energy and Industrial Technology Development Organization (NEDO) under project JPNP26007.
  T.O. was supported by JSPS KAKENHI Grant Number JP25K03176.
  Y.K. was supported by JST PRESTO, Japan, Grant Number JPMJPR2612.
  K.O. was supported by JST SPRING, Grant Number JPMJSP2108.
  N.S. was supported by JST BOOST, Japan Grant Number JPMJBS2418, and the Teijin Scholarship Foundation.

\bibliographystyle{IEEEtran}
\bibliography{root}

\begin{thebibliography}{10}
\providecommand{\url}[1]{#1}
\csname url@rmstyle\endcsname
\providecommand{\newblock}{\relax}
\providecommand{\bibinfo}[2]{#2}
\providecommand\BIBentrySTDinterwordspacing{\spaceskip=0pt\relax}
\providecommand\BIBentryALTinterwordstretchfactor{4}
\providecommand\BIBentryALTinterwordspacing{\spaceskip=\fontdimen2\font plus
\BIBentryALTinterwordstretchfactor\fontdimen3\font minus
  \fontdimen4\font\relax}
\providecommand\BIBforeignlanguage[2]{{%
\expandafter\ifx\csname l@#1\endcsname\relax
\typeout{** WARNING: IEEEtran.bst: No hyphenation pattern has been}%
\typeout{** loaded for the language `#1'. Using the pattern for}%
\typeout{** the default language instead.}%
\else
\language=\csname l@#1\endcsname
\fi
#2}}

\bibitem{pi05}
K.~Black \emph{et~al.}, ``$\pi_{0.5}$: a vision-language-action model with
  open-world generalization,'' in \emph{Proceedings of The 9th Conference on
  Robot Learning}, ser. Proceedings of Machine Learning Research, vol.
  305.\hskip 1em plus 0.5em minus 0.4em\relax PMLR, 2025, pp. 17--40.

\bibitem{nvidia2025gr00tn1openfoundation}
NVIDIA \emph{et~al.}, ``{GR00T N1}: An open foundation model for generalist
  humanoid robots,'' arXiv:2503.14734 [cs.RO], 2025.

\bibitem{NEURIPS2025_300ccb21}
K.~Black, M.~Galliker, and S.~Levine, ``Real-time execution of action chunking
  flow policies,'' in \emph{Advances in Neural Information Processing Systems},
  vol.~38, 2025, pp. 33\,383--33\,407.

\bibitem{paraschos2013promp}
A.~Paraschos, C.~Daniel, J.~R. Peters, and G.~Neumann, ``Probabilistic movement
  primitives,'' in \emph{Advances in Neural Information Processing Systems},
  vol.~26, 2013.

\bibitem{zhao2023act}
T.~Z. Zhao, V.~Kumar, S.~Levine, and C.~Finn, ``Learning fine-grained bimanual
  manipulation with low-cost hardware,'' in \emph{Proceedings of Robotics:
  Science and Systems}, Daegu, Republic of Korea, July 2023.

\bibitem{11127493}
S.~Dasari, O.~Mees, S.~Zhao, M.~K. Srirama, and S.~Levine, ``The ingredients
  for robotic diffusion transformers,'' in \emph{2025 IEEE International
  Conference on Robotics and Automation (ICRA)}, 2025, pp. 15\,617--15\,625.

\bibitem{Osa18a}
T.~Osa, J.~Pajarinen, G.~Neumann, J.~A. Bagnell, P.~Abbeel, and J.~Peters, ``An
  algorithmic perspective on imitation learning,'' \emph{Foundations and Trends
  in Robotics}, vol.~7, no. 1-2, pp. 1--179, 2018.

\bibitem{doi:10.1177/02783649261417694}
T.~Tsuji, Y.~Kato, G.~Solak, H.~Zhang, T.~Petrič, F.~Nori, and A.~Ajoudani,
  ``A survey on imitation learning for contact-rich tasks in robotics,''
  \emph{The International Journal of Robotics Research}, 2026.

\bibitem{Kingma14}
D.~P. Kingma and M.~Welling, ``Auto-encoding variational {Bayes},'' in
  \emph{Proceedings of the International Conference on Learning Representations
  (ICLR)}, 2014.

\bibitem{Sohn15}
K.~Sohn, H.~Lee, and X.~Yan, ``Learning structured output representation using
  deep conditional generative models,'' in \emph{Advances in Neural Information
  Processing Systems}, vol.~28, 2015.

\bibitem{chi2023diffusionpolicy}
C.~Chi, S.~Feng, Y.~Du, Z.~Xu, E.~Cousineau, B.~C. Burchfiel, and S.~Song,
  ``Diffusion policy: Visuomotor policy learning via action diffusion,'' in
  \emph{Proceedings of Robotics: Science and Systems}, Daegu, Republic of
  Korea, July 2023.

\bibitem{9561138}
S.~Mysore, B.~Mabsout, R.~Mancuso, and K.~Saenko, ``Regularizing action
  policies for smooth control with reinforcement learning,'' in \emph{2021 IEEE
  International Conference on Robotics and Automation (ICRA)}, 2021, pp.
  1810--1816.

\bibitem{Ijspeert2001}
A.~Ijspeert, J.~Nakanishi, and S.~Schaal, ``Trajectory formation for imitation
  with nonlinear dynamical systems,'' in \emph{Proceedings of the IEEE/RSJ
  International Conference on Intelligent Robots and Systems (IROS)}, vol.~2,
  2001, pp. 752--757.

\bibitem{6797340}
A.~J. Ijspeert, J.~Nakanishi, H.~Hoffmann, P.~Pastor, and S.~Schaal,
  ``Dynamical movement primitives: Learning attractor models for motor
  behaviors,'' \emph{Neural Computation}, vol.~25, no.~2, pp. 328--373, 2013.

\bibitem{li2023prodmp}
G.~Li, Z.~Jin, M.~Volpp, F.~Otto, R.~Lioutikov, and G.~Neumann, ``{ProDMP}: A
  unified perspective on dynamic and probabilistic movement primitives,''
  \emph{IEEE Robotics and Automation Letters}, vol.~8, no.~4, pp. 2325--2332,
  2023.

\bibitem{beast2025}
H.~Zhou, W.~Liao, X.~Huang, Y.~Tang, F.~Otto, X.~Jia, X.~Jiang, S.~Hilber,
  G.~Li, Q.~Wang, {\"O}.~E. Ya{\u{g}}murlu, N.~Blank, M.~Reuss, and
  R.~Lioutikov, ``{BEAST}: Efficient tokenization of {B}-splines encoded action
  sequences for imitation learning,'' in \emph{Advances in Neural Information
  Processing Systems}, 2025.

\bibitem{10480552}
P.~M. Scheikl, N.~Schreiber, C.~Haas, N.~Freymuth, G.~Neumann, R.~Lioutikov,
  and F.~Mathis-Ullrich, ``Movement primitive diffusion: Learning gentle
  robotic manipulation of deformable objects,'' \emph{IEEE Robotics and
  Automation Letters}, vol.~9, no.~6, pp. 5338--5345, 2024.

\bibitem{ho2020ddpm}
J.~Ho, A.~Jain, and P.~Abbeel, ``Denoising diffusion probabilistic models,'' in
  \emph{Advances in Neural Information Processing Systems}, vol.~33, 2020, pp.
  6840--6851.

\end{thebibliography}

\end{document}